\documentclass[proc]{edpsmath}

\usepackage{macrosBonus}
\usepackage{amssymb,url}
\usepackage{todonotes}

\begin{document}
	
	\title{Learning the Distribution with Largest Mean: \\ Two Bandit Frameworks}\thanks{This work was partially supported by the CIMI (Centre International de Math\'ematiques et d'{In\-for\-ma\-tique}) Excellence program while Emilie Kaufmann visited Toulouse in November 2015.
		The authors acknowledge the support of the French Agence Nationale de la Recherche (ANR), under grants ANR-13-BS01-0005 (project SPADRO) and ANR-13-CORD-0020 (project ALICIA).}%
	\author{Emilie Kaufmann}\address{CNRS \& Univ. Lille, UMR 9189 (CRIStAL), Inria Lille Nord-Europe (SequeL), \texttt{emilie.kaufmann@univ-lille1.fr}}
	\author{Aur\'elien Garivier}\address{Institut de Math\'ematiques de Toulouse, UMR 5219, Universit\'e de Toulouse, CNRS, \texttt{aurelien.garivier@math.univ-toulouse.fr}}
	%
	%
	\begin{abstract} Over the past few years, the multi-armed bandit model has become increasingly popular in the machine learning community, partly because of applications including online content optimization. This paper reviews two different sequential learning tasks that have been considered in the bandit literature ; they can be formulated as (sequentially) learning which distribution has the highest mean among a set of distributions, with some constraints on the learning process. For both of them (\emph{regret minimization} and \emph{best arm identification}) we present recent, asymptotically optimal algorithms. We compare the behaviors of the sampling rule of each algorithm as well as the complexity terms associated to each problem.\end{abstract}
	\begin{resume} Le mod\`ele stochastique dit de bandit \`a plusieurs bras soul\`eve ces derni\`eres ann\'ees un grand int\'er\^et dans la communaut\'e de l'apprentissage automatique, du fait notamment de ses applications \`a l'optimisation de contenu sur le web. Cet article pr\'esente deux probl\`emes d'apprentissage s\'equentiel dans le cadre d'un mod\`ele de bandit qui peuvent \^etre formul\'es comme la d\'ecouverte de la distribution ayant la moyenne la plus \'elev\'ee dans un ensemble de distributions, avec certaines contraintes sur le processus d'apprentissage. Pour ces deux objectifs (\emph{minimisation du regret} d'une part et \emph{identification du meilleur bras} d'autre part), nous pr\'esentons des algorithmes optimaux, en un sens asymptotique. Nous comparons les strat\'egies d'\'echantillonnage employ\'ees par ces deux types d'algorithmes ainsi que les quantit\'es caract\'erisant la complexit\'e de chacun des probl\`emes. \end{resume}
	\maketitle
	\section*{Introduction}
	
	Bandit models can be traced back to the 1930s and the work of \cite{Thompson33} in the context of medical trials. It addresses the idealized situation where, for a given symptom, a doctor has several treatments at her disposal, but has no prior knowledge about their efficacies. These efficacies need to be \emph{learnt} by allocating treatments to patients and observing the result. As the doctor aims at healing as many patients as possible, she would like to select the best treatment as often as possible, even though it is unknown to her at the beginning. After each patient, the doctor takes the outcome of the treatment into account in order to decide which treatment to assign to the next patient: the learning process is \emph{sequential}. 
	
	This archetypal situation is mathematically captured by the \emph{multi-armed bandit model}. It involves an \emph{agent} (the doctor) interacting with a set of $K$ probability distribution $\nu^1,\dots,\nu^K$ called \emph{arms} (the treatments), which she may sequentially sample. The mean of arm $a$  (which is unknown to the agent) is denoted by $\mu_a$. At round $t$, the agent selects an arm $A_t\in\{1,\dots,K\}$ and subsequently observes a sample $X_t \sim \nu^{A_t}$ from the associated distribution. The arm $A_t$ is selected according to a \emph{sampling strategy} denoted by $\pi=(\pi_t)_{t\geq 1}$, where $\pi_t$  maps the history of past arm choices and observations $A_1,X_1,\dots,A_{t-1},X_{t-1}$ to an arm. In a simplistic model for the clinical trial example, each arm is a Bernoulli distribution that indicates the success or failure of the treatment. After sampling an arm (giving a treatment) at time $t$, the doctor observes whether the patient was healed ($X_t=1$) or not ($X_t=0$). In this example as in many others, the samples gathered can be considered as \emph{rewards}, and a natural goal for the agent is to adjust her sampling strategy so as to maximize the expected sum  $\bE[\sum_{t=1}^T X_t]$ of the rewards gathered up to some given \emph{horizon} $T$. This is equivalent to minimizing the \emph{regret}
	\[\mathrm{R}^\pi(T) = \mu^*T - \bE\left[\sum_{t=1}^T X_t\right]\;,\]
	which is defined as the gap between the expected efficiency of the strategy $\pi$ and the expected cumulated reward of an \emph{oracle strategy} always playing the best arm $a^* = \argmax{a} \ \mu_a$ that has mean $\mu^* = \max_a \mu_a$.
	
	A sampling strategy minimizing  the regret should not only learn which arm has the highest mean: it should also not incur too big losses during this learning phase. In other words, it has to achieve a good trade-off between \emph{exploration} (experimenting all the arms in order to estimate their means) and \emph{exploitation} (focusing on the arm that appears best so far). Despite its simplicity, the multi-armed bandit model already captures the fundamental dilemma inherent to \emph{reinforcement learning}~\cite{SuttonBarto98}, where the goal is to learn how to act optimally in a random environment based on numeric feedback. The fundamental model of reinforcement learning is the Markov Decision Process \cite{Puterman94MDP}, which involves the additional notion of system state; a bandit model is simply a Markov Decision Process with a single state. 
	
	Sometimes, rewards actually correspond to profits for the agent. In fact, the imaginatively named \emph{multi-armed bandits} refer to casino slot machines: a player sequentially selects one of them (also called a \emph{one-armed bandit}), draws its arm, and possibly collects her wins. While the model was initially motivated by clinical trials, modern applications involve neither bandits, nor casinos, but for example the design of recommender systems \cite{LinUCB11}, or more generally content optimization. Indeed, a bandit algorithm may be used by a company for dynamically selecting which version of its website to display to each  user, in order to maximize the number of \emph{conversions} (purchase or subscription for example). In the case of two competing options, this problem is known as  \emph{A/B testing}. It motivates the consideration of a different optimization problem in a bandit model: rather than continuously changing its website, the company may prefer to experiment during a \emph{testing phase} only, which is aimed at identifying the best version, and then to use that one consistently for a much bigger audience.
	
	In such a testing phase, the objective is different: one aims at learning which arm has highest mean \emph{without constraint} on the cumulative reward. In other words, the company agrees to lose some profit during the testing phase, as long as the length of this phase is as short as possible. In this framework, called \emph{best arm identification}, the sampling rule is designed so as to identify the arm with highest mean as fast and as confidently as possible. Two alternative frameworks are considered in the literature. In the \emph{fixed-budget setting} \cite{Bubeck10BestArm}, the length of the trial phase is given and the goal is to minimize the probability of misidentifying the best arm. In the \emph{fixed-confidence setting} \cite{EvenDaral06}, a risk parameter $\delta$ is given and the procedure is allowed to choose when the testing phase stops. It must guarantee that the misidentification probability is smaller than $\delta$ while minimizing the \emph{sample complexity}, that is the  expected number of samples required before electing the arm. Although the study of best arm identification problems is relatively recent in the bandit literature, similar questions were already addressed in the 1950s under the name \emph{ranking and identification} problems \cite{Bechofer:54,Bechofer:al68}, and they are also related to the sequential adaptive hypothesis testing framework introduced by \cite{Chernoff59}.   
	
	In this paper, we review a few algorithms for both regret minimization and best arm identification in the fixed-confidence setting. The algorithms and results are presented for simple classes of parametric bandit models, and we explain along the way how some of them can be extended to more general models. In each case we introduce an asymptotic notion of optimality and present algorithms that are asymptotically optimal. Our optimality notion is \emph{instance-dependent}, in the sense that we characterize the minimal regret or minimal sample complexity achievable on each specific bandit instance. The paper is structured as follows: we introduce in Section~\ref{sec:Tools} the parametric bandit models considered in the paper, and present some useful probabilistic tools for the analysis of bandit algorithms. We discuss the regret minimization problem in Section~\ref{sec:RegretMinimization} and the best arm identification problem in Section~\ref{sec:BestArmIdentification}. We comment in Section~\ref{sec:Discussion} on the different behaviors of the algorithms aimed at these distinct objectives, and on the different information-theoretic quantities characterizing their complexities.

	\section{Parametric Bandit Models and Useful Tools} \label{sec:Tools}
	
	\subsection{Some Assumptions on the Arms Distributions} \label{subsec:Model}
	
	Unless specified otherwise, we assume in the rest of the paper that all the arms belong to a class of distributions parameterized by their means, $\cD_{I} = \{ \nu_{\mu} : \mu \in I\}$, where $I$ is an interval of $\R$. We assume that for all $\mu \in I$, $\nu_{\mu}$ has a density denoted by $f_{\mu}$ with respect to some fixed reference measure, and that $\bE_{X \sim \nu_\mu}[X] = \mu$. For all $(\mu,\mu') \in I^2$ we introduce
	\[d(\mu,\mu') := \mathrm{KL}(\nu_{\mu},\nu_{\mu'}) = \bE_{X \sim \nu_\mu}\left[\log \frac{f_{\mu}(X)}{f_{\mu'}(X)}\right],\]
	the Kullback-Leibler divergence between the distribution of mean $\mu$ and that of mean $\mu'$. We shall in particular consider examples in which $\cD_{I}$ forms a one-parameter exponential family (e.g. Bernoulli distributions, Gaussian distributions with known variance, exponential distributions), for which there is a closed form formula for the divergence function $d$ (see, e.g. \cite{KLUCBJournal}). 
	
	Under this assumption, a bandit model is fully described by a vector $\bm \mu = (\mu_1,\dots,\mu_K)$ in $I^K$ such that $\nu^{a} = \nu_{\mu_a}$ for all $a \in \{1,\dots,K\}$.  We denote by $\bP_{\bm \mu}$ and $\bE_{\bm \mu}$ the probability and expectation under the bandit model $\bm \mu$.
	Under $\bP_{\bm \mu}$, the sequence $(Y_{a,s})_{s\in \N^*}$ of successive observations from arm $a$ is i.i.d. with law $\nu_{\mu_a}$, and the families $(Y_{a,s})_a$  are independent. Given a strategy $\pi$, we let $N_a^\pi(t) = \sum_{s=1}^t\ind_{(A_s=a)}$ be the number of draws of arm $a$ up to and including round $t\geq 1$. Hence, upon selection of the arm $A_t$, the observation made at round $t$ is $X_t = Y_{A_t,N_{A_t}^\pi(t)}$. When the strategy $\pi$ is clear from the context, we may remove the superscript $\pi$ and write simply $N_a(t)$. We define $\hat{\mu}_{a,s}= \frac{1}{s}\sum_{i=1}^s Y_{a,i}$ as the empirical mean of the first $s$ observations from arm $a$, and $\hat{\mu}_a(t) = \hat{\mu}_{a,N_a(t)}$ as the empirical mean of arm $a$ at round $t$ of the bandit algorithm.

	For the two frameworks that we consider, regret minimization and best arm identification, we adopt the same approach. First, we propose a lower bound on the target quantity (regret or sample complexity). Then, we propose strategies whose regret or sample complexity asymptotically matches the lower bound. Two central tools to derive lower bounds and algorithms are \emph{changes of distributions} and \emph{confidence intervals}.

	\subsection{Changes Of Distribution}\label{subsec:CD}
	
	Problem-dependent lower bounds in the bandit literature all rely in the end on change of distribution arguments (see e.g. \cite{LaiRobbins85bandits,BurnKat96,MannorTsi04,Bubeck10BestArm}). In order to control the probability of some event under the bandit model $\bm \mu$, the idea is to consider an alternative bandit model $\bm \lambda$ under which some assumptions on the strategy make it is easier to control the probability of this event. This alternative model $\bm\lambda$ should be \emph{close} enough to $\bm\mu$, in the sense that the \emph{transportation cost} should  not be too high. This transportation cost is related to the log-likelihood ratio of the observations up to time $t$, that we denote by 
	\[L_t^\pi(\bm\mu,\bm \lambda) = \sum_{a=1}^K \sum_{s=1}^{N_a^\pi(t)} \log \frac{f_{\mu_a}(Y_{a,s})}{f_{\lambda_a}(Y_{a,s})}.\]
	Letting $\cF_t = \sigma ( X_1,\dots, X_t)$ be the $\sigma$-field generated by the observations up to time $t$, it is indeed well known that for all $\cE \in \cF_t$, $\bP_{\bm\mu}(\cE) = \bE_{\bm \lambda}\left[\ind_{\cE} \exp\left(L_t^\pi(\bm \mu,\bm\lambda)\right)\right]$.
	
	The most simple way of writing changes of distribution (see~\cite{COLT14,Combes14Unimodal} and \cite{GMS16}) directly relates the \emph{expected} log-likelihood ratio of the observations under two bandit models to the probability of any event under the two models. If $S$ is a stopping time, one can show that for any two bandit models $\bm \mu, \bm \lambda$ and for any event in $\cF_{S}$,
	\[\bE_{\bm \mu}[L_S^\pi (\bm \mu,\bm \lambda)] \geq \kl(\bP_{\bm \mu}(\cE),\bP_{\bm \lambda}(\cE)),\]
	where $\kl(x,y) = x \log(x/y) + (1-x)\log((1-x)/(1-y))$ is the binary relative entropy, i.e. the Kullback-Leibler divergence between two Bernoulli distributions of means $x$ and $y$.
	Using Wald's lemma, one can show that in the particular case of bandit models, the expected log-likelihood ratio can be expressed in terms of the expected number of draws of each arms, which yields the following result. 
	
	\begin{lemma}\label{lem:CD} Let $S$ be a stopping time. For any event $\cE \in \cF_{S}$,
		\[\sum_{a=1}^K\bE_{\bm\mu}[N_a^\pi(S)]d(\mu_a,\lambda_a) \geq \kl(\bP_{\bm\mu}(\cE),\bP_{\bm\lambda}(\cE))\;.\]
	\end{lemma}
	Two different proofs of this result can be found in \cite{JMLR15} and \cite{GMS16}, in which a slightly more general result is derived based on the entropy contraction principle. As we will see in the next sections, this lemma is particularly powerful to prove lower bounds on the regret or the sample complexity, as both quantities are closely related to the expected number of draws of each arm. 
	
	
	\subsection{Confidence Intervals}\label{subsec:CI}
	
	In both the regret minimization and best arm identification frameworks, the sampling rule has to decide which arm to sample from at a current round, based on the observations gathered at previous rounds. This decision may be based on the set of statistically plausible values for the mean of each arm $a$, that is materialized by a \emph{confidence interval} on $\mu_a$. Note that  in this sequential learning framework, this interval has to be built based on a \emph{random number of observations}. 
	
	The line of research leading to the UCB1 algorithm \cite{Aueral02} worked under the assumption that each arm is a bounded distribution supported in $[0,1]$. Bounded distributions are particular examples of sub-Gaussian distributions. A random variable $X$ is said to be $\sigma^2$-sub-Gaussian if $\bE[e^{\lambda (X - \bE[X])}] \leq \exp({\lambda^2\sigma^2}/{2})$ holds for all $\lambda\in\R$. Hoeffding's lemma states that distributions with a support bounded in $[a,b]$ are $(b-a)^2/4$-sub-Gaussian. If arm $a$ is $\sigma^2$-sub-Gaussian, Hoeffding's inequality together with a union bound to handle the random number of observations permits to show that 
	\begin{equation}\bP\left(\hat{\mu}_a(t) + \sqrt{\frac{2\sigma^2 \gamma}{N_a(t)}} < \mu_a\right) \leq t e^{-\gamma}.\label{bound:UCB1}\end{equation}
	Hence on can build an upper-confidence bound on $\mu_a$ with probability of coverage $1 - \delta$ by setting $\gamma = \log(t/\delta)$.
	
	There are two levels of improvement here. First, under more specific assumption on the arms (for example if the arms belong to some exponential family of distributions), Chernoff's inequality has an explicit form that can be used directly in place of Hoeffding's inequality. It states that $\bP\left(\hat{\mu}_{a,s} > x\right)\leq \exp(-sd(x,\mu_a))$, where $d(x,y)$ is the KL-divergence function defined in Section~\ref{subsec:Model}. Then, to handle the random number of observations, a peeling argument can be used rather than a union bound. This argument, initially developed in the context of Markov order estimation (see~\cite{GL11}), was used in~\cite{GarivierMoulines11,Bubeck:Thesis} under sub-gaussian assumption. Combining these two ideas \cite{AOKLUCB} show that, letting 
	\[u_a(t) = \sup \{q : N_a(t)\, d (\hat{\mu}_a(t) , q) \leq \gamma \}\]
	one has 
	\begin{equation}\bP\left(u_a(t) < \mu_a\right) \leq e \lceil\gamma \log(t)\rceil e^{-\gamma}.\label{bound:KLUCB}\end{equation}
	
	The improvement can be measured by specifying this result to Bernoulli distributions, for which the two bounds \eqref{bound:UCB1} and \eqref{bound:KLUCB} hold. By Pinsker's inequality $\kl(x,y) > 2(x-y)^2$, it holds that $u_a(t) \leq \hat{\mu}_a(t) + \sqrt{2\sigma^2 \gamma/N_a(t)}$. Hence for $\gamma$ and $t$ such that $e \lceil \gamma \log(t) \rceil \leq t$, $u_a(t)$ is a smaller upper-confidence bound on $\mu_a$ with the same coverage guarantees. As we will see in the next sections, such refined confidence intervals have yield huge improvements in the bandit literature, and lead to simple UCB-type algorithms that are asymptotically optimal for regret minimization.

	\section{Optimal Strategies For Regret Minimization}\label{sec:RegretMinimization}
	
	After the initial work of \cite{Thompson33}, bandit models were studied again in the 1950s, with for example the paper of \cite{Robbins52Freq}, in which the notion of regret is introduced. Interestingly, a large part of the early work on bandit models takes a slightly different \emph{Bayesian} perspective: the goal is also to maximize the expected sum of rewards, but the expectation is also computed over a prior distribution for the arms (see \cite{Berry:Fristedt85} for a survey). It turns out that this Bayesian multi-armed bandit problem can be solved exactly using dynamic programming \cite{Bellman:Bay56}, but the exact solution is in most cases intractable. Practical solutions may be found when one aims at maximizing the sum of \emph{discounted} rewards over an infinite horizon: the seminal paper of \cite{Gittins79} shows that the Bayesian optimal policy has a simple form where, at each round an index is computed for each arm and the arm with highest index is selected.   
	
	Gittins's work motivated the focus on \emph{index policies}, where an index is computed for each arm as a selection procedure. Such index policies have also emerged in the ``frequentist'' literature on multi-armed bandits. Some asymptotic expansions of the index put forward by Gittins  were proposed. They have led to new policies that could be studied directly, forgetting about their Bayesian roots. This line of research includes in particular the seminal work of \cite{LaiRobbins85bandits}.

	\subsection{A Lower Bound on the Regret}
	
	In 1985, Lai and Robbins characterized the optimal regret rate in one-parameter bandit models, by providing an asymptotic lower bound on the regret and a first index policy with a matching regret~\cite{LaiRobbins85bandits}. In order to understand this lower bound, one can  first observe that the regret can be expressed  in terms of the number of draws of each sub-optimal arm. Indeed, a simple conditioning shows that for any strategy $\pi$, 
	\begin{equation}
	R^{\pi}_{\bm \mu}(T) = \bE\left[\sum_{t=1}^T (\mu^* - \mu_{A_t})\right] = \sum_{a : \mu_a < \mu^*} (\mu^* - \mu_a)\bE_{\bm \mu}\left[N_a^\pi(T)\right],\label{def:regNbDraws}
	\end{equation}
	where we recall that $N_a^\pi(t) = \sum_{s=1}^t \ind_{(A_s =t)}$ is the number of times arm $a$ has been selected up to time $t$. A strategy is said to be \emph{uniformly efficient} if its regret is small on every bandit model in our class, that is if for all $\bm \mu \in I^K$ and for every $\alpha \in ]0,1],$ $R_{\bm \mu}^\pi(T) = o(T^\alpha)$. 
	
	\begin{theorem}{\cite{LaiRobbins85bandits}} \label{thm:LBLR} Any uniformly efficient strategy $\pi$ satisfies, for all $\bm \mu \in I^K$, 
		\[
		\forall a : \mu_a < \mu^*, \ \ \liminf_{T\rightarrow \infty} \frac{\bE_{\bm \mu}[N_a^\pi(T)]}{\log(T)} \geq \frac{1}{d(\mu_a,\mu^*)}\;.
		\]
	\end{theorem}
	
	By Equation~\eqref{def:regNbDraws}, this result directly provides a logarithmic regret lower bound on the regret: 
	\begin{equation}\label{eq:LBRegret}\liminf_{T\rightarrow \infty}\frac{R^{\pi}_{\bm\mu}(T)}{\log(T)} \geq \sum_{a : \mu_a < \mu^*} \frac{\mu^* - \mu_a}{d(\mu_a,\mu^*)} \stackrel{\mathrm{def}}{=} C(\bm \mu)\;.\end{equation}
	This lower bound motivates the definition of an \emph{asymptotically optimal algorithm} (on a set of parametric bandit models $\cD_I$) as an algorithm for with for all $\bm \mu \in I^K$, the regret is asymptotically upper bounded by  $C(\bm \mu)\log(T)$. This defines an \emph{instance-dependent} notion of optimality, as we want an algorithm that attains the best regret rate for every bandit instance $\bm \mu$. However, for some instances  $\bm\mu$ such that some arms are very close to the optimal arm, the constant $C(\bm \mu)$ may be really large and the $C(\bm \mu)\log(T)$ bound is not very interesting in finite-time. For such instances, one may prefer having regret upper bounds that scale in $\sqrt{KT}$ and are independent of $\bm \mu$, matching the \emph{minimax} regret lower bound obtained by \cite{PLG06,Bubeck:Survey12} for Bernoulli bandits:
	\[\inf_{\pi} \sup_{\bm \mu \in I} R^{\pi}_{\bm \mu}(T) \geq \frac{1}{20}\sqrt{KT}.\]
	
	Logarithmic instance-dependent regret lower bound have also been obtained under more general assumptions for the arms distributions \cite{BurnKat96}, and even in some examples of structured bandit models, in which they take a less explicit form \cite{GravesLai97,Combes14Lip}. All these lower bounds rely on a change of distribution argument, and we now explain how to easily obtain the lower bound of Theorem~\ref{thm:LBLR} by using the tool described in Section~\ref{subsec:CD}, Lemma~\ref{lem:CD}. 
	
	Fixing a suboptimal arm $a$ in the bandit model $\bm \mu$, we define an alternative bandit model $\bm\lambda$ such that $\lambda_i = \mu_i$ for all $i\neq a$ and $\lambda_a = \mu^* + \epsilon$. In $\bm \lambda$, arm $a$ is now the optimal arm, hence a uniformly efficient algorithm will draw this arm very often. As arm $a$ is the only arm that has been modified in $\bm \lambda$, the statement in Lemma~\ref{lem:CD} takes the simple form: 
	\[\bE_{\bm \mu}[N_a^\pi(T)] d(\mu_a,\mu^* + \epsilon) \geq \kl\left(\bP_{\bm \mu}(A_T),\bP_{\bm \lambda}(A_T)\right),\]
	for any event $A_T \in \cF_T$. Now the event $A_T := \left(N_a(T) < T/2\right)$ is very likely under $\bm \mu$ in which $a$ is sub-optimal, and very unlikely under $\bm \lambda$ in which $a$ is optimal. More precisely, the uniformly efficient assumption permits to show that $\bP_{\bm \mu}(A_T) \rightarrow 1$ and $\bP_{\bm \lambda}(A_T) \leq {o(T^\alpha)}/{T}$ for all $\alpha$ when $T$ goes to infinity. This leads to $\kl\left(\bP_{\bm \mu}(A_T),\bP_{\bm \lambda}(A_T)\right) \sim \log(T)$ and proves Theorem~\ref{thm:LBLR}.  
	
	\subsection{Asymptotically Optimal Index Policies and Upper Confidence Bounds}
	
	Lai and Robbins also proposed the first algorithm whose regret matches the lower bound \eqref{eq:LBRegret} and this first asymptotically optimal algorithm is actually an index policy, i.e. it is of the form  
	\[A_{t+1} = \argmax{a \in \{1,\dots,K\}} \ U_a(t),\]
	but the proposed indices $U_a(t)$ are quite complex to compute. \cite{Agrawal:95,KatRob:95Gauss} later proposed slightly more simple indices and show that they can be interpreted as \emph{Upper Confidence Bounds} (UCB) on the unknown means of the arms. UCB-type algorithms were popularized by \cite{Aueral02}, who introduce the UCB1 algorithm for (non-parametric) bandit models with bounded rewards, and give the first \emph{finite-time} upper bound on its regret. Simple indices like those of UCB1 can be used more generally for $\sigma^2$-sub-Gaussian rewards, and take the form 
	\[U_a(t) = \hat{\mu}_a(t) + \sqrt{\frac{2\sigma^2f(t)}{N_a(t)}},\]
	for some function $f$ which controls the confidence level. While the original choice of \cite{Aueral02} is too conservative, one may safely choose  $f(t)=\log(t)$ in practice; obtaining finite-time regret bounds is somewhat easier with a slightly larger choice, as in~\cite{AOKLUCB}. With such a choice, for Bernoulli distributions (which are $1/4$-sub-Gaussian), the regret of this index policy can be shown to be   
	\[\sum_{a : \mu_a < \mu^*}^K \frac{1}{2(\mu^*-\mu_a)} \log(T) + O(\sqrt{\log(T)}),\]
	which is only order-optimal with respect to the lower bound \eqref{eq:LBRegret}, as by Pinsker inequality $d(\mu_a,\mu^*) > 2(\mu^*-\mu_a)^2$. Since the work of \cite{Aueral02}, several improvements of UCB1 have been proposed. They aimed at providing finite-time regret guaranteess that would match the asymptotic lower bound~\eqref{eq:LBRegret} (see the review~\cite{Bubeck:Survey12}). Among them, the kl-UCB algorithm studied by \cite{KLUCBJournal} is shown to be asymptotically optimal when the arms belong to a one-parameter exponential family. This algorithm is an index policy associated with 
	\[u_a(t) = \max \big\{ q : N_a(t)\, d(\hat{\mu}_a(t), q) \leq f(t) \big\},\]
	for the same choice of an exploration function $f$ as mentioned above. The discussion of Section~\ref{subsec:CI} explains why this index is actually an upper confidence bound on $\mu_a$: choosing $f(t) = \log(t) + 3\log\log(t),$ one has $\bP_{\bm \mu}\left(u_a(t) \geq \mu_a\right) \gtrsim 1 - 1/\big[t\log^{2}(t)\big]$. For this particular choice, \cite{KLUCBJournal} give a finite-time analysis of kl-UCB, proving its asymptotic optimality. To conclude on UCB algorithms, let us mention that several improvements have been proposed. A simple but significant one is obtained by replacing $f(t)$ by $\log(t/N_a(t))$ in the definition of $u_a(t)$, leading to a variant sometimes termed kl-UCB$^+$ which has a slightly better empirical performance, but also minimax guarantees that plain UCB algorithms do not enjoy (for a discussion and related ideas, see the OCUCB algorithm of~\cite{L16}, \cite{MG17} and the references therein). 
	\subsection{Beyond the Optimism Principle}
	
\begin{figure}[h]
		\centering
		\includegraphics[width=0.49\linewidth]{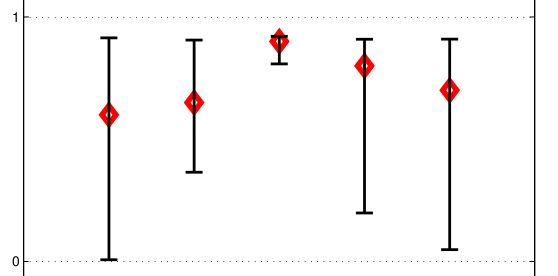}
		\includegraphics[width=0.49\linewidth]{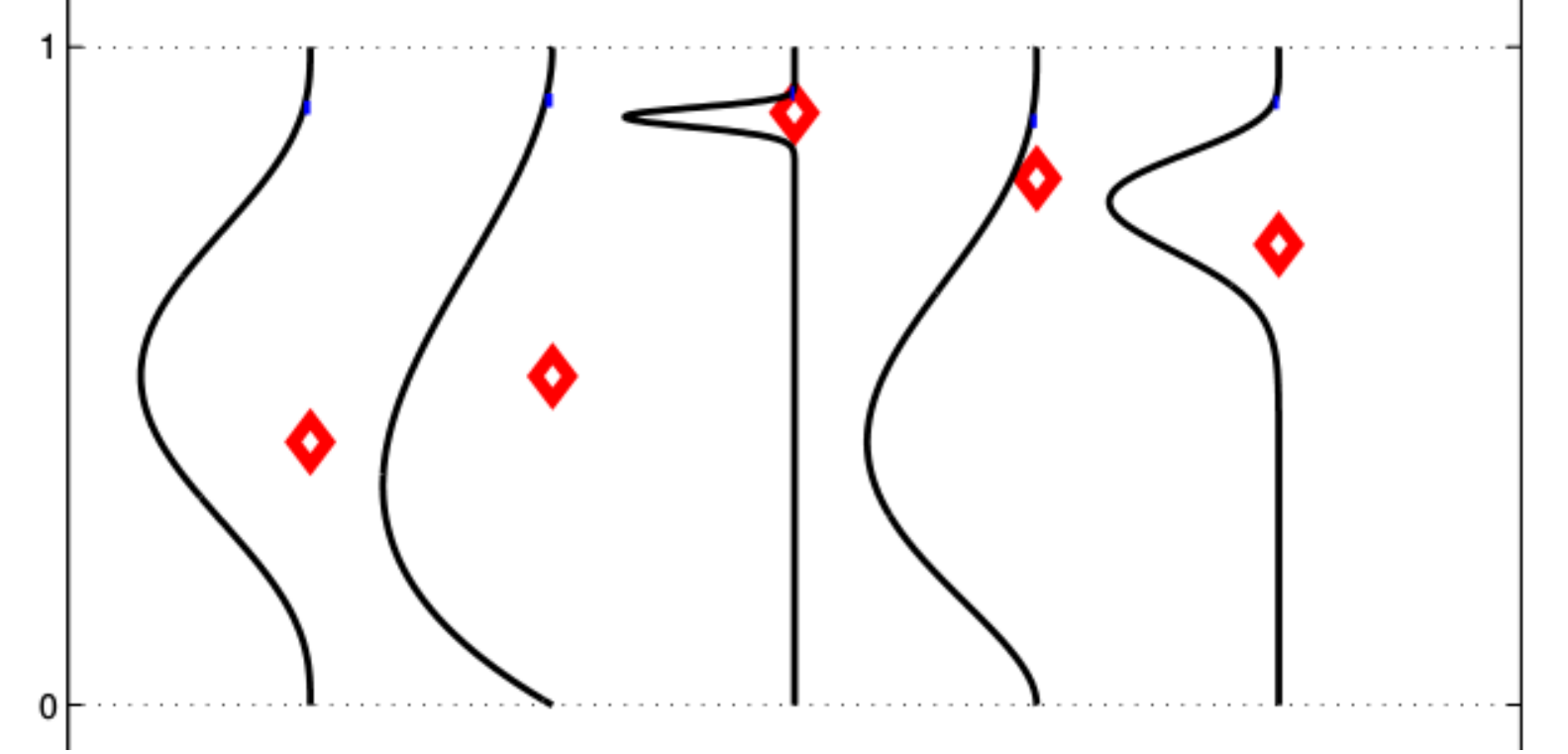}
		\caption{\label{fig:IllustrationAlgos} Frequentist versus Bayesian algorithms in a five-armed Bernoulli bandit model (the red diamonds are the unknown means). On the left, the confidence intervals used by kl-UCB, on the right the Beta posterior distributions used by Bayesian algorithms.}
	\end{figure}

	For simple parametric bandit models, in particular when rewards belong to a one-parameter exponential family, we showed that the regret minimization problem is solved, at least in an asymptotic sense: the kl-UCB algorithm, for example, attains the best possible regret rate on every problem instance. All the UCB-type algorithms described in the previous section are based on the so called principle of ``optimism in face of uncertainty''. Indeed, at each round of a UCB algorithm the confidence intervals materialize the set of bandit models that are compatible with our observations (see Figure~\ref{fig:IllustrationAlgos}, left), and choosing the arm with largest UCB amounts to acting optimally in an ``optimistic'' model in which the mean of each arm would be equal to its best possible value. This optimism principle has also been successfully applied in some structured bandit models \cite{YadLinear11}, as well as in reinforcement learning \cite{Auer:UCRL10} and other related problems~\cite{BG13}. 
	
	While the Lai and Robbins' lower bound provides a good guideline to design algorithms, it has sometimes been misunderstood as a justification of the wrong \emph{folk theorem} which is well-known by practitioners mostly interested in using bandit algorithms: ``no strategy can have a regret smaller than $C({\bm \mu})\log(t)$, which is reached by good strategies".
	But experiments often infirm this claim: it is easy to show settings and algorithms where the regret is much smaller than $C({\bm \mu})\log(t)$ and does not look like a logarithmic curve. The reason is twofold: first, Lai and Robbins' lower result is asymptotic; a close look at its proof shows that it is relevant only when the horizon $T$ is so large that any reasonable policy has identified the best arm with high probability; second, it only states that the regret \emph{divided by $\log(t)$} cannot always be smaller than $C({\bm \mu})$. In~\cite{GKL16}, a more simple but similar bandit model of complexity $C'({\bm \mu})$ is given where some strategy is proved to have a regret smaller than $C'({\bm \mu})\log(t)-c\log(\log(t))$ for some positive constant $c$.  
	
	Some recent works try to complement this result and to give a better description of what can be observed in practice. Notably, \cite{GMS16} focuses mainly on the initial regime: the authors show in particular that all strategies suffer a \emph{linear regret} before $T$ reaches some problem-dependent value. When the problem is very difficult (for example when the number of arms is very large) this initial phase may be the only observable one... They give non-asymptotic inequalities, and above all show a way to prove lower bounds which may lead to further new results (see e.g.~\cite{GKL16}).  
	It would be of great interest (but technically difficult) to exhibit an intermediate regime where, after this first phase, statistical estimation becomes possible but is still not trivial. This would in particular permit to discriminate from  a theoretical perspective between all the bandit algorithms that are now known to be asymptotically optimal, but for which significant differences may be observed in practice.
	
	Indeed, one drawback of the kl-UCB algorithm is the need to construct tight confidence intervals (as explained in Section~\ref{subsec:CI}), which may not be generalized easily beyond simple parametric models. More flexible, Bayesian algorithms have recently also been shown to be asymptotically optimal, and to have good empirical performances. 
	Given a prior distribution one the arms, Bayesian algorithms are simple procedures exploiting the \emph{posterior distributions} of each arm. In the Bernoulli example, assuming a uniform prior on the mean of each arm, the posterior distribution of $\mu_a$ at round $t$, defined as the conditional distribution of $\mu_a$ given past observations, is easily seen to be a Beta distribution with parameters given by the number of ones and zeros observed so far from the arm. A Bayesian algorithm uses the different posterior distributions, that are illustrated in Figure~\ref{fig:IllustrationAlgos} (right), to choose the next arm to sample from. 
	
	The Bayes-UCB algorithm of \cite{AISTATS12} exploits these posterior distributions in an optimistic way: it selects at round $t$ the arm whose posterior on the mean has the largest quantile of order $1-1/t$. Another popular algorithm, Thompson Sampling, departs from the optimism principle by selecting arms at random according to their probability of being optimal. This principle was proposed by \cite{Thompson33} as the very first bandit algorithm, and can easily be implemented by drawing one sample from the posterior distribution of the mean of each arm, and selecting the arm with highest sample. This algorithm, also called probability matching, was rediscovered in the 2000s for its good empirical performances in complex bandit models \cite{Scott10,LiChapelle11}, but its first regret analysis dates back to \cite{AGCOLT12}. Both Thompson Sampling and Bayes-UCB have been shown recently to be asymptotically optimal in one-parameter models, for some choices of the prior distribution \cite{ALT12,AGAISTAT13,AOS16}. These algorithms are also quite generic, as they can be implemented in any bandit model in which one can define a prior distribution on the arms, and draw samples from the associated posterior. For example, they can be used in (generalized) linear bandit models, that can model recommendation tasks where the features of the items are taken into account (see, e.g. \cite{AGAISTAT13} and the Chapter 4 of \cite{MaThese}).

	\section{Optimal Strategies for Best Arm Identification}\label{sec:BestArmIdentification}
	Finding the arm with largest mean (without trying to maximize the \emph{cumulated} rewards) is quite a different task and relates more to classical statistics. It can indeed be cast into the framework of \emph{sequential adaptive hypothesis testing} introduced by \cite{Chernoff59}. In this framework, one has to decide which of the (composite) hypotheses 
	\[H_1 : ``\mu_1 > \max_{i\neq 1} \mu_i"\,, \ H_2 = ``\mu_2 > \max_{i\neq 2} \mu_i" \,, \dots, \ \ H_K = ``\mu_K > \max_{i\neq K} \mu_i"\]
	is true. In order to gain information, one can select at each round one out of $K$ possible \emph{experiments}, each of them consisting in sampling from one of the marginal distributions (arms). Moreover, one has to choose when to stop the trial and decide for one of the hypotheses. Rephrased in a ``bandit'' terminology, a strategy consists of
	\begin{itemize}
		\item a \emph{sampling rule}, that specifies which arm $A_t$ is selected at time $t$ ($A_t$ is $\cF_{t-1}$-measurable), 
		\item a \emph{stopping rule} $\tau$, that indicates when the trial ends ($\tau$ is a stopping time wrt $\cF_t$),
		\item a \emph{recommendation rule} $\hat{a}_{\tau}$ that provides, upon stopping, a guess for the best arm ($\hat{a}_{\tau}$ is $\cF_\tau$-measurable).
	\end{itemize}
	
	However, the objective of the fixed-confidence best arm identification problem differs from that of \cite{Chernoff59}, where one aims at minimizing a risk measure of the form 
	\[R(\bm \mu) = r_{a^*(\bm \mu)}\bP_{\bm \mu}(\hat{a}_{\tau} \neq a^*(\bm \mu)) + c\bE_{\bm\mu}[\tau],\]
	where $r_i$ is the cost of wrongly rejecting hypothesis $H_i$ and $c$ is a cost for sampling. Modern bandit literature rather focuses on so-called \emph{$(\epsilon,\delta)$-PAC strategies} (for Probably Approximately Correct) which output, with high probability, an arm whose mean is within $\epsilon$ of the mean of the best arm:
	\[\forall \bm \mu \in I^K, \ \ \bP_{\bm \mu}\left(\mu_{\hat{a}_\tau} > \mu^* - \epsilon\right) \geq 1 - \delta\;.\]
	The goal is to build a $(\epsilon,\delta)$-PAC strategy with a sample complexity $\bE_{\bm \mu}[\tau]$ that is as small as possible. 
	For simplicity, we focus here on the case $\epsilon=0$: a strategy is called $\delta$-PAC\footnote{It would be more correct to call it a \emph{$\delta$-PC (Probably Correct) strategy}.} if 
	\[\forall \bm \mu \in \cS, \ \ \bP_{\bm \mu}\left(\hat{a}_{\tau} = a^*(\bm \mu)\right) \geq 1 - \delta,\]
	where $\cS = \{\bm \mu \in I^K : \exists i : \mu_i > \max_{j\neq i} \mu_j\}$ is the set of bandit models that have a unique optimal arm.
	
	We show in the next section that, as in the regret minimization framework, there exists an instance-dependent lower bound on the sample complexity of any $\delta$-PAC algorithm.
	We further present an algorithm whose sample complexity matches the lower bound, at least in the asymptotic regime where $\delta$ goes to $0$. 
	It is remarkable that this optimal algorithm, described in Section~\ref{sec:TaS}, is actually a by-product of the lower bound analysis described in Section~\ref{sec:BAILB}, which sheds light on how a good strategy should distribute the draws between the arms.
	
	\subsection{The Sample Complexity of $\delta$-PAC Best Arm Identification}\label{sec:BAILB}
	
	The first lower bound on the sample complexity of a $(\epsilon,\delta)$-PAC algorithm was given by \cite{MannorTsi04}.
	Particularized to the case $\epsilon=0$, the lower bound says that for Bernoulli bandit models with means in $[0,\alpha]$, there exists a constant $C_{\alpha}$ and a subset of the sub-optimal arms $\cK_\alpha$ such that for any $\delta$-PAC algorithm
	\[\bE_{\bm \mu}[\tau] \geq C_{\alpha}\left[ \sum_{a \in \cK_\alpha} \frac{1}{(\mu^* - \mu_a)^2}\right]\log \left(\frac{1}{8\delta}\right).\]
	Following this result, the literature has provided several $\delta$-PAC strategies together with upper bounds on their sample complexity, mostly under the assumption that the rewards are bounded in $[0,1]$. Existing strategies fall into two categories: those based on successive eliminations \cite{EvenDaral06,Karnin:al13}, and those based on confidence intervals \cite{Shivaramal12,Gabillon:al12,Jamiesonal14LILUCB}. 
	For all these algorithms, under a bandit instance such that $\mu_1 > \mu_2 \geq \dots \geq \mu_K$, the number of samples used can be shown to be of order
	\[C\left[\frac{1}{(\mu_1 - \mu_2)^2} + \sum_{a=2}^K\frac{1}{(\mu_1-\mu_a)^2}\right]\log\left(\frac{1}{\delta}\right) + o_{\delta \rightarrow 0}\left(\log \frac{1}{\delta}\right),\]
	where $C$ is a (large) numerical constant. While explicit finite-time bounds on $\tau$ can be extracted from most of the papers listed above, we mostly care here about the first-order term in $\delta$, when $\delta$ goes to zero. Both the upper and lower bounds take the form of a sum over the arms of an individual complexity term (involving the inverse squared gap with the best or second best arm), but there is a gap as those sums do not involve the same number of terms; in addition, loose  multiplicative constants make it hard to identify the exact minimal sample complexity of the problem.  
	
	As for the regret minimization problem, the true sample complexity can be expected to involve information-theoretic quantities (like the Kullback-Leibler divergence between arms distributions), for which the quantities above appear to be only surrogates; for example, for Bernoulli distributions, it holds that $2(\mu_1 - \mu_a)^2< d(\mu_a,\mu_1) < (\mu_1 - \mu_a)^2/(\mu_1(1-\mu_1))$. For exponential families, it has been shown that incorporating the KL-based confidence bounds described in Section~\ref{subsec:CI} into existing algorithms lowers the sample complexity \cite{COLT13} but the true sample complexity was only recently obtained by \cite{GK16}. The result, and its proof, are remarkably simple:
	
	\begin{theorem}\label{thm:LBPAC} Let $\bm \mu \in \cS$, define $\Alt(\bm \mu) := \{ \bm \lambda \in \cS : a^*(\bm \lambda) \neq a^*(\bm \mu)\}$ and let  $\Sigma_K = \big\{ w \in [0,1]^K : \sum_{a=1}^K w_a =1\big\}$ be the set of probability vectors. Any $\delta$-PAC algorithm satisfies 
		\[\bE_{\bm \mu}[\tau] \geq T^*(\bm \mu)\, \kl(\delta,1-\delta)\;,\]
		where
		\[T^*(\bm \mu)^{-1} = \sup_{w \in \Sigma_K} \inf_{\bm \lambda \in \Alt(\bm \mu)} \sum_{a=1}^K w_a d (\mu_a,\lambda_a)\;.\]
	\end{theorem}
	
	This lower bound against relies on a change of distribution, but unlike Lai and Robbins' result (and previous results for best arm identification), it is not sufficient to individually lower bound the expected number of draws of each arm using a single alternative model. One needs to consider the set $\Alt(\bm\mu) = \{\bm \lambda : a^*(\bm \lambda)\neq a^*(\bm\mu)\}$ of all possible alternative models $\bm \lambda$ in which the optimal arm is different from the optimal arm of $\bm\mu$.
	
	Given a $\delta$-PAC algorithm, let $\cE = (\hat{a}_\tau \neq \hat{a}^*(\bm \mu))$.
	For \emph{any} $\bm\lambda \in \Alt(\bm\mu)$, the $\delta$-PAC property implies that $\bP_{\bm\mu}(\cE) \leq \delta$ while $\bP_{\bm \lambda}(\cE) \geq 1-\delta$. Hence, by Lemma~\ref{lem:CD},
	\[\sum_{a=1}^K \bE_{\bm\mu}[N_a(\tau)] d(\mu_a,\lambda_a) \geq \kl(\delta,1-\delta).\]
	Combining \emph{all} the inequalities thus obtained for the different possible values of $\bm\lambda \in \Alt(\bm\mu)$, we conclude that:
	\begin{align*}
		\kl(\delta,1-\delta) &\leq  \inf_{\bm \lambda \in \Alt(\bm\mu)}\sum_{a=1}^K \bE_{\bm\mu}[N_a(\tau)] d(\mu_a,\lambda_a)\\
		&\leq 
		\bE_{\bm\mu}[\tau]\left(\inf_{\bm \lambda \in \Alt(\bm\mu)}\sum_{a=1}^K \frac{\bE_{\bm\mu}[N_a(\tau)]}{\bE_{\bm\mu}[\tau]} d(\mu_a,\lambda_a)\right)\\
		&\leq 
		\bE_{\bm\mu}[\tau]\left(\sup_{w \in \Sigma_K}\inf_{\bm \lambda \in \Alt(\bm\mu)}\sum_{a=1}^K w_a d(\mu_a,\lambda_a)\right) \;.
	\end{align*}
	In the last step, we use the fact that the vector $(\bE_{\bm\mu}[N_a(\tau)]/\bE_{\bm\mu}[\tau])$ sums to one: upper bounding by the worst probability vector $w$ yields a bound that is independent of the algorithm.
	
	We thus obtain the (not fully explicit, but simple) lower bound of Theorem~\ref{thm:LBPAC} that holds under the parametric assumption of Section~\ref{subsec:Model}. Its form involving an optimization problem is reminiscent of the early work of \cite{Agrawaletal89LBGene,GravesLai97} that provide a lower bound on the regret in general, possibly structured bandit models. For best arm identification, \cite{Rajesh15Oddball} consider the particular case of Poisson distribution in which there is only one arm that is different from the others, where a very nice formula can be derived for the sample complexity. For general exponential family bandit models, we now provide a slightly more explicit expression of $T^*(\bm\mu)$, that permits to efficiently compute it.

	\subsection{An Asymptotically Optimal Algorithm}\label{sec:TaS}
	
	\subsubsection{Computing the complexity and the optimal weights}
	The proof of Theorem~\ref{thm:LBPAC} reveals that the quantity 
	\[w^*(\bm \mu) \in \argmax{w \in \Sigma_K} \ \inf_{\bm \lambda \in \Alt(\bm \mu)} \sum_{a=1}^K w_a d (\mu_a,\lambda_a) \]
	can be interpreted as a vector of \emph{optimal proportions}, in the sense that any strategy matching the lower bound should satisfy $\bE_{\bm \mu}[N_a(\tau)]/\bE_{\bm\mu}[\tau] \simeq w^*_a(\bm \mu)$. Some algebra shows that the above optimization problem has a unique solution, and provides an efficient way of computing $w^*(\bm\mu)$ for any $\bm\mu$, which boils down to numerically solving a series of scalar equations. In this section, we shall assume that $\bm \mu$ is such that $\mu_1 > \mu_2 \geq \dots \geq \mu_K$.
	
	First, when the distribution belong to a one-dimensional exponential family (which we assume in the rest of this section), one can solve the inner optimization over $\bm\lambda$ in closed form, using Lagrange duality. This yields:
	\[w^*(\bm \mu) \in \argmax{w \in \Sigma_K} \ \min_{a\neq 1} \left[w_1 d \left(\mu_1,\frac{w_1\mu_1 + w_a\mu_a}{w_1+w_a}\right) + w_a d \left(\mu_a,\frac{w_1\mu_1 + w_a\mu_a}{w_1+w_a}\right)\right].\]
	Then, one can prove that at the optimum the $K-1$ quantities in the $\min$ are equal. Introducing their common value as an auxiliary variable, one can show that the computation of $w^*(\bm\mu)$ reduces to solving a one-dimensional optimization problem. For all $a\neq 1$, one introduces the strictly increasing mapping 
	\begin{eqnarray*}
		g_a : [0,+\infty[ & \longrightarrow & [0,d(\mu_1,\mu_a)[ \\
		x & \mapsto & d\left(\mu_1,\frac{\mu_1 + x \mu_a}{1+x}\right)+ xd\left(\mu_a,\frac{\mu_1 + x \mu_a}{1+x}\right),
	\end{eqnarray*}
	and defines $x_a : [0,d(\mu_1,\mu_a)[ \rightarrow [0,+\infty[$ to be its inverse mapping. With this notation, the following Lemma~\ref{lem:ExplicitForm} provides a way to compute $w^*(\bm\mu)$. The scalar equation $F_{\bm \mu}(y)=0$ defined therein may be solved using binary search. At each step of the search, the solution $x_a(y)$ of the equation $g_a(x)=y$ can again be computed by using binary search, or by Newton's method. This algorithm is available as a \texttt{julia} code at \url{https://github.com/jsfunc/best-arm-identification}.

	\begin{lemma}\label{lem:ExplicitForm} For every $a\in\{1,\dots,K\}$, 
		\begin{equation}\label{eq:nuofx}
		w^*_a(\bm \mu) = \frac{x_a(y^*)}{\sum_{a=1}^Kx_a(y^*)}\;,\end{equation}
		where $y^*$ is the unique solution of the equation $F_{\bm\mu}(y)=1$, where
		\begin{equation}\label{eq:relation} F_{\bm\mu}:y\mapsto \sum_{a=2}^K \frac{d\left(\mu_1,\frac{\mu_1 + x_a(y)\mu_a}{1+x_a(y)}\right)}{d\left(\mu_a,\frac{\mu_1 + x_a(y)\mu_a}{1+x_a(y)}\right)} \end{equation}
		is a continuous, increasing function on $[0,d(\mu_1,\mu_2)[$ such that $F_{\bm\mu}(0)=0$ and $F_{\bm\mu}(y)\to \infty$ when $y\to d(\mu_1,\mu_2)$.
	\end{lemma}
	
	This result yields an efficient algorithm for computing $T^*(\bm\mu)$. But can a closed form formula be derived, at least in some special cases?
	In the two-armed case, it is easy to see that $T^*(\bm\mu)$ is equal to the inverse Chernoff information between the two arms (see~\cite{COLT14}). However, no closed form is available when $K\geq 3$, even for simple families of distributions. For Gaussian arms with known variance $\sigma^2$, only the following bound is known which captures $T^*(\bm\mu)$ up to a factor $2$:
	\[\frac{2\sigma^2}{(\mu_1-\mu_2)^2}+\sum_{a=2}^K \frac{2\sigma^2}{(\mu_1-\mu_a)^2} \leq T^*(\bm \mu) \leq 2 \left[\frac{2\sigma^2}{(\mu_1-\mu_2)^2}+\sum_{a=2}^K \frac{2\sigma^2}{(\mu_1-\mu_a)^2}\right].\]
	Note that $T^*(\bm \mu)$ may be much smaller than ${4\sigma^2 K}/{(\mu_1-\mu_2)^2}$, which is the minimal number of samples required by a strategy using uniform sampling (for which $N_a(t)/t \simeq 1/K$). An optimal strategy actually uses quite unbalanced weights $w^*(\bm \mu)$. 
	
	\subsubsection{An algorithm inspired from the lower bound}
	Back to general exponential families, building on the lower bound and our ability to compute $w^*(\bm\mu)$, we now introduce an efficient algorithm whose sample complexity matches the lower bound, at least for small values of $\delta$. This Track-and-Stop strategy consists of two elements: 
	\begin{itemize}
		\item a \emph{tracking sampling rule}, that forces the proportion of draws of each arm $a$ to converge to the associated optimal proportion $w_a^*(\bm\mu)$, by using the plug-in estimates $w^*(\hat{\bm\mu}(t))$,
		\item the \emph{Chernoff stopping rule}, that can be interpreted as the stopping rule of a sequential Generalized Likelihood Ratio Test (GLRT), whose closed form in this particular problem is very similar to our lower bound. 
	\end{itemize}
	When stopping, our guess $\hat{a}_\tau$ is the empirical best arm. We now describe the sampling and stopping rule in details before presenting the theoretical guarantees for Track-and-Stop. 
	
	\subsubsection{The Tracking sampling rule}
	
	Let $\hat{\bm\mu}(t) = (\hat{\mu}_1(t), \dots, \hat{\mu}_K(t))$ be the current maximum likelihood estimate of $\bm\mu$ at time $t$: \[\hat{\mu}_a(t)=\frac{1}{N_a(t)}\sum_{s\leq t}X_s \ind\{A_s=a\}\;.\]
	A first idea for matching the proportions $w^*(\bm \mu)$ is to track the plug-in estimates $w^*(\hat{\bm\mu}(t))$, by drawing at round $t$ the arm $a$ whose empirical proportion of draws lags furthest behind the estimated target  $w_a^*(\hat{\bm\mu}(t))$. But a closer inspection shows that (sufficiently fast) convergence of $\hat{\bm\mu}(t)$ towards the true parameter $\bm\mu$ requires some ``forced exploration'' to make sure each arm has not been under-sampled. More formally, defining $F_t = \{a : N_a(t) < \sqrt{t} - K/2\}$, the Tracking rule is defined as 
	\[A_{t+1} \in \left\{\begin{array}{ll}
	\underset{a \in F_t}{\text{argmin}}  \ N_a(t) \ \text{if} \ F_t \neq \emptyset & (\textit{forced  exploration})\\
	\underset{1 \leq a \leq K}{\text{argmax}} \left\{w_a^*(\hat{\bm\mu}(t)) -N_a(t)/t\right\} \ \text{otherwise} & (\textit{tracking the plug-in estimate})
	\end{array}
	\right.\]
	Simple combinatorial arguments prove that the Tracking rule draws each arm at least $(\sqrt{t} -K/2)_+ -1$ times at round $t$, and relate the gap between $N_a(t)/t$ and $w_a^*(\bm\mu)$ to the gap between $w_a^*(\bm\mu)$ and $w_a(\hat{\bm\mu}(t))$. This permits in particular to show that the Tracking rule has the following desired behavior:
	
	\begin{proposition}\label{prop:ConvergenceFraction} The Tracking sampling rule satisfies
		\[\bP_{\bm\mu}\left(\lim_{t \rightarrow \infty} \frac{N_a(t)}{t} = w_a^*(\bm \mu)\right) = 1\;.\] 
	\end{proposition}
	
	\subsubsection{The Chernoff stopping rule}
	Let $\ell(X_1,\dots,X_t ; \bm\lambda)$ be the log-likelihood of the observations up to time $t$ under a bandit model $\bm\lambda$ and define 
	\begin{eqnarray}
	\hat{Z}(t) & = & \frac{\sup_{\bm \lambda}\ell(X_1,\dots,X_t ; \bm \lambda)}{\sup_{\bm \lambda \in \Alt(\hat{\bm \mu}(t))} \ell(X_1,\dots,X_t ; \bm \lambda)} = \frac{\ell(X_1,\dots,X_t ; \bm \hat{\bm\mu}(t))}{\sup_{\bm \lambda \in \Alt(\hat{\bm \mu}(t))} \ell(X_1,\dots,X_t ; \bm \lambda)}.
	\nonumber\end{eqnarray} 
	Intuitively, this generalized likelihood ratio $\hat{Z}(t)$ is large if the current maximum-likelihood estimate is far apart from its ``closest alternative'' $\tilde{\bm\mu}(t)$ defined as the parameter maximizing the likelihood under the constraint that it belongs to $\Alt(\hat{\bm\mu}(t))$, i.e. that its optimal arm is different from that of $\hat{\bm\mu}(t)$. This idea can be traced back to the work of \cite{Chernoff59}, in which $\hat{Z}(t)$ is interpreted as the Neyman-Pearson statistic for testing the (data-dependent) pseudo-hyptothesis $``\bm \mu = \hat{\bm\mu}(t)"$ against $``\bm \mu = \tilde{\bm \mu}(t)"$, based on all samples available up to round $t$. The analysis of Chernoff, however, only applies to two discrete hypotheses, whereas the best arm identification problem requires to consider $K$ continuous hypotheses.
	
	In the paper \cite{GK16}, we provide new insights on this Chernoff stopping rule, formally defined as 
	\begin{equation}\tau_{\delta} = \inf \left\{ t \in \N : \hat{Z}(t) > \beta(t,\delta) \right\}\;,\label{def:Stopping}\end{equation}
	where $\beta(t,\delta)$ is some threshold function. 
	
	The first problem is to set the threshold $\beta(t,\delta)$ such that the probability of error of Track-and-Stop is upper bounded by $\delta$. Our analysis relies on expressing $\hat{Z}(t)$ in terms of pairwise sequential GLRTs of $``\mu_a < \mu_b"$ against $``\mu_a \geq \mu_b"$, for which we provide tight bounds on the type I error. Indeed, letting 
	\[\hat{Z}_{a,b}(t) := \log\frac{\max_{\lambda_a\geq \lambda_b} p_{\lambda_a}\left(\underline{X}^a_{N_a(t)}\right)p_{\lambda_b}\left(\underline{X}^b_{N_b(t)}\right)}{\max_{\lambda_a\leq \lambda_b} p_{\lambda_a}\left(\underline{X}^a_{N_a(t)}\right)p_{\lambda_b}\left(\underline{X}^b_{N_b(t)}\right)}\;,\]
	where $\underline{X}^a_{N_a(t)} = (X_s: A_s=a, s\leq t)$ is a vector that contains the observations of arm $a$ available at time $t$, and where $p_{\lambda}(V_1,\dots,V_n)$ is the likelihood of $n$ i.i.d. observations drawn from $\nu_{\lambda}$, one can show that  
	\begin{equation}\hat{Z}(t) = \min_{b \neq \hat{a}(t)} \hat{Z}_{\hat{a}(t),b}(t)\;,\label{def:ChernoffStopAlt}\end{equation}
	where $\hat{a}(t)$ is the empirical best arm at round $t$. In other words, one stops when for each arm $b$ that is different from the empirical best arm $\hat{a}(t)$, a GLRT would reject the (data-dependent) pseudo-hypothesis $``\mu_{\hat{a}(t)} < \mu_b"$. This expression also allows for a simple computation of $\hat{Z}(t)$, as  $\hat{\mu}_a(t) > \hat{\mu}_b(t)$ implies
	\[\hat{Z}_{a,b}(t) = N_a(t) d\left(\hat{\mu}_a(t),\frac{N_a(t)\hat{\mu}_a(t) + N_b(t)\hat{\mu}_b(t)}{N_a(t)+N_b(t)}\right) 
	+ N_b(t) d\left(\hat{\mu}_b(t),\frac{N_a(t)\hat{\mu}_a(t) + N_b(t)\hat{\mu}_b(t)}{N_a(t)+N_b(t)}\right).\]
	Under the Tracking sampling rule, it is easy to see that the $\hat{Z}(t)$ grows linearly with $t$, hence with a $\beta(t,\delta)$ that is sub-linear in $t$, $\tau_\delta$ is also surely finite. The probability of error of the Chernoff sampling rule is thus upper bounded by 
	\[\bP_{\bm \mu}\left(\hat{a}_\tau \neq 1\right) \leq \bP_{\bm \mu}\left(\exists t \in \N : \hat{a}(t) \neq 1 , \hat{Z}(t) > \beta(t,\delta)\right)\leq \sum_{a=2}^K\bP_{\bm \mu}\left(\exists t \in \N : \hat{Z}_{a,1}(t) > \beta(t,\delta)\right).\]
	In the Bernoulli case, the following lemma permits to prove that the Chernoff stopping rule is $\delta$-PAC for the choice \begin{equation}\beta(t,\delta) = \log\left(\frac{2(K-1)t}{\delta}\right).\label{ChoiceBeta}\end{equation}
	
	\begin{lemma}\label{lem:deltaPAC} 
		Let $\delta \in (0,1)$. For any sampling strategy, if $\mu_a<\mu_b$, 
		$\bP\left(\exists t\in \N : \hat{Z}_{a,b}(t) > \log(2t/\delta)\right) \leq \delta$.
	\end{lemma}
	
	Another rewriting of the Chernoff stopping rule permits to understand why it achieves the optimal sample complexity when coupled to the Tracking stopping rule. Indeed, using the particular form of the likelihood in an exponential family yields 
	\[\hat{Z}(t) = t \left(\inf_{\bm\lambda \in \Alt(\hat{\bm\mu}(t))} \sum_{a=1}^K \frac{N_a(t)}{t}\; d\big(\hat{\mu}_a(t),\lambda_a\big)\right).\]
	This expression is reminiscent of the lower bound of Theorem~\ref{thm:LBPAC}. When $t$ is large one expects $\hat{\bm\mu}(t)$ to be close to $\bm \mu$ thanks to the forced exploration, and $N_a(t)/t$ to be close to $w^*_a(\bm\mu)$, due to Proposition~\ref{prop:ConvergenceFraction}. Hence one has $\hat{Z}(t) \simeq t/T^*(\bm \mu)$ for large values of $t$. Thus, with the threshold function \eqref{ChoiceBeta}, for small $\delta$, $\tau_\delta$ is asymptotically upper bounded by the smallest $t$ such that $t \geq T^*(\bm\mu)\log(2(K-1)t/\delta)$, which is of order $T^*(\bm \mu) \log(1/\delta)$ for small values of $\delta$.

	\subsubsection{Optimality of Track-and-Stop}
	
	In the previous section, we sketched an upper bound on the number of samples used by Track-and-Stop that holds with probability one in the Bernoulli case. \cite{GK16} also propose an asymptotic upper bound on the expected sample complexity of this algorithm, beyond the Bernoulli case. The results are summarized below. 
	
	\begin{theorem}\label{thm:optimTaS} Let $\alpha > 1$. There exists a constant $C=C_{\alpha,K}$ such that for all $\delta \in ]0,1[$ the Track-and-Stop strategy with threshold \[\beta(t,\delta) = \log \left(\frac{C t^\alpha}{\delta}\right)\]
		is $\delta$-PAC and satisfies 
		\[\limsup_{\delta \rightarrow 0}\frac{\bE_{\bm\mu}[\tau_\delta]}{\log(1/\delta)} \leq \alpha T^*(\bm\mu).\]
		In the Bernoulli case, one can set $\alpha=1$ and $C=2(K-1)$.
	\end{theorem}
	
	Hence, Track-and-Stop can be qualified as asymptotically optimal, in the sense that its sample complexity matches the lower bound of Theorem~\ref{thm:LBPAC}, when $\delta$ tends to zero. Inspired by the regret minimization study, an important direction of future work is to obtain finite-time upper bounds on the sample complexity of an algorithm, that would still asymptotically match the lower bound of Theorem~\ref{thm:LBPAC}. A different line of research has studied, for sub-Gaussian rewards, the asymptotic behavior of the sample complexity for fixed values of $\delta$ in a regime in which the gap between the best and second best arm goes to zero \cite{Jamiesonal14LILUCB,Chen16OptimalAlt}, leading to a different notion of optimality. Hence, we should aim for the best of both worlds: an algorithm with a finite-time sample complexity upper bound that would also match the lower bound obtained in this alternative asymptotic regime. 
	
	Finally, while Theorem~\ref{thm:optimTaS} gives asymptotic results for Track-and-Stop, we would like to highlight the practical impact of this algorithm. Experiments in \cite{GK16} reveal that for relatively ``large'' values of $\delta$ (e.g. $\delta=0.1$) the sample complexity of Track-and-Stop appears to be twice smaller than that of state-of-the-art algorithms in generic scenarios. The sampling rule of Track-and-Stop is slightly more computationally demanding than that of its competitors, as it requires to compute $w^*(\hat{\bm\mu}(t))$ at each round. However, the Tracking sampling rule is the most naive idea, and we will investigate whether other simple heuristics could be used to guarantee that the empirical proportions of draws converge towards the optimal proportions $w^*(\bm \mu)$, while being amenable for finite-time analysis. 
	
	\section{Discussion}\label{sec:Discussion}
	
	It is known at least since \cite{Bubeckal11} that good algorithms for regret minimization and pure-exploration are expected to be different: small regrets after $t$ time steps imply a large probability of error $\bP_{\bm \mu}(\mu^* \neq \mu_{\hat{a}(t)})$, where $\hat{a}(t)$ is the recommendation for the best arm at time $t$. In the dual fixed confidence setting that was studied in this paper, we provided other elements to assess the difference of the regret minimization and best arm identification problems.
	
\begin{figure}[h]
		\includegraphics[width=0.47\textwidth]{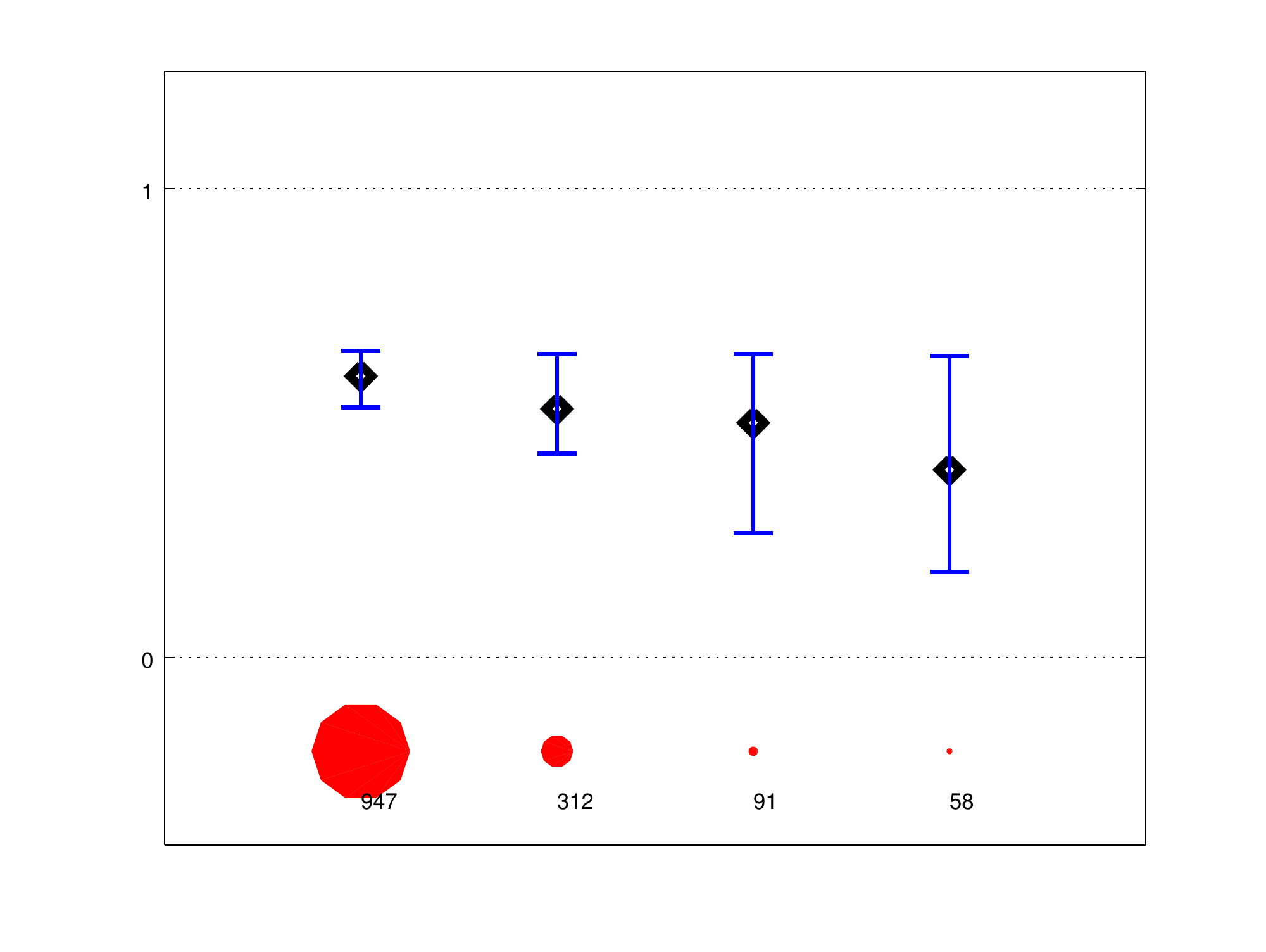}
		\includegraphics[width=0.47\textwidth]{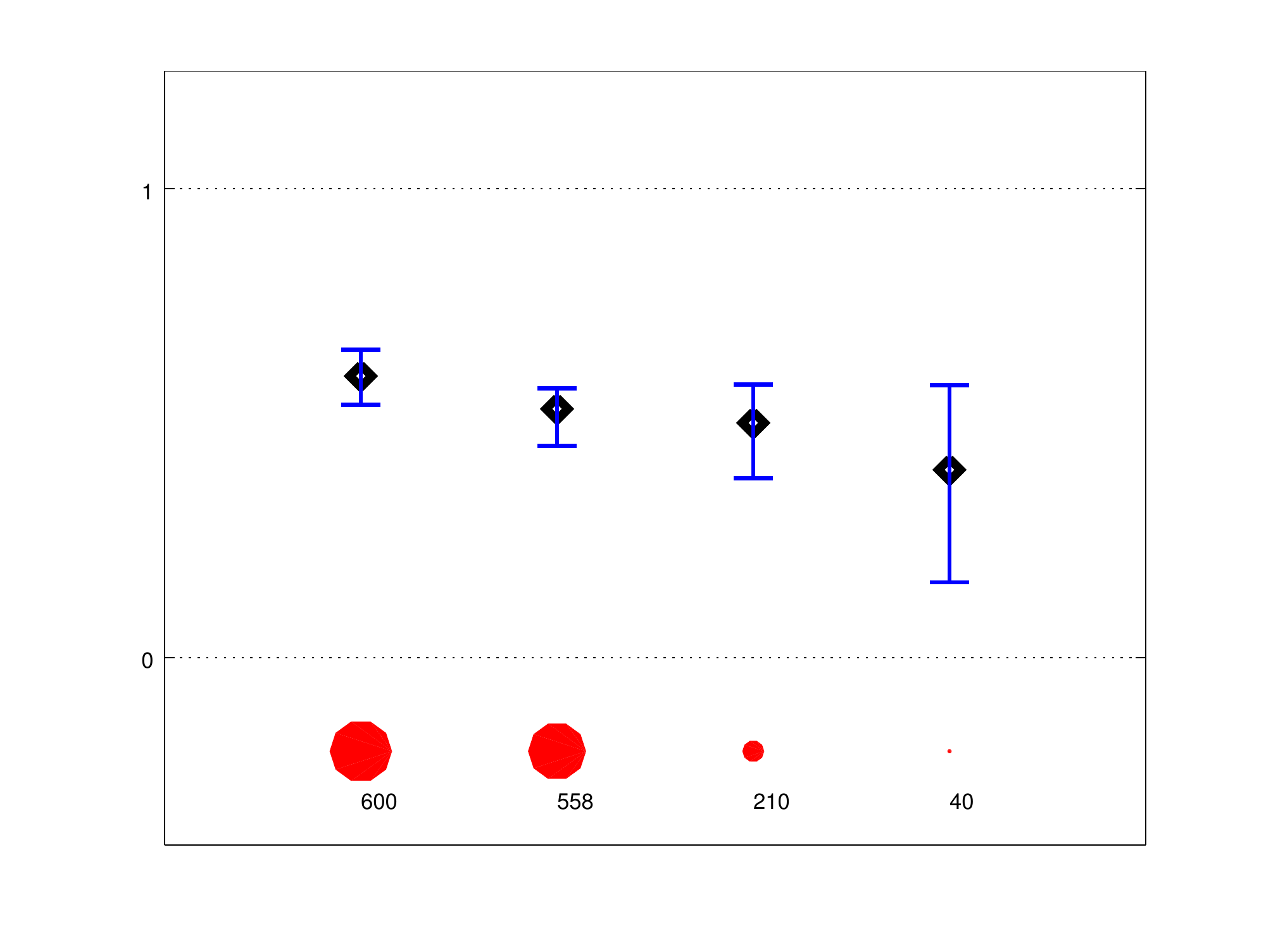}
		\caption{\label{fig:DifferenceBAIRM} Number of draws of the kl-UCB (left) and Track-and-Stop (right) algorithms on one run of length $t=1408$ (the stopping time of Track-and-Stop). Confidence intervals on the unknown mean of each arm (black diamond) are also displayed.}
	\end{figure}

	First, the sampling strategy used by both type of algorithms are very different. Regret minimizing algorithm draw the best arm most of the time ($t - O(\log(t))$ times in $t$ rounds) while each sub-optimal arm gathers a vanishing proportion of draws. On the contrary, identifying the best arm requires the proportions of arm draws to converge to a vector $w^*(\bm \mu)$ will all non-zero components. Figure~\ref{fig:DifferenceBAIRM} illustrates this different behavior: the number of draws of each arm and associated KL-based confidence intervals are displayed for the kl-UCB (left) and Track-and-Stop (right) strategies. As expected, Track-and-Stop  draws more frequently than its competitor the close-to-optimal arms, and has therefore tighter confidence intervals on their means. 
	
	We also emphasize that the information-theoretic quantities characterizing the complexity of the two problems are different. For regret minimization, we saw that the minimal regret of uniformly efficient (u.e.) strategies satisfies:
	\[\inf_{\pi \in \{u.e.\}} \lim_{T\rightarrow \infty} \frac{R_{\bm \mu}^\pi(T)}{\log(T)} = \sum_{a : \mu_a < \mu^*}\frac{\mu^* - \mu_a}{d(\mu_a,\mu^*)},\]
	where $d(x,y)$ is the Kullback-Leibler divergence between the distribution of mean $x$ and the distribution of mean $y$ in our class. For best arm identification,
	\[\inf_{\pi \in \{PAC\}} \lim_{\delta\rightarrow 0} \frac{\bE_{\bm\mu}[\tau_\delta]}{\log(1/\delta)} = T^*(\bm \mu),\]
	where $T^*(\bm \mu)$ is the solution of an optimization problem expressed with Kullback-Leibler divergences between arms that has no closed form solution for more than two arms.
	
	Although regret minimization and best arm identification are two very different objectives, both in terms of algorithms and of complexity, best arm identification tools have been used within regret minimization algorithms in so-called \emph{Explore-Then-Commit} strategies \cite{PerchetRigollet13Covariates,Perchet15Batched}. For minimizing regret up to a horizon $T$ such strategies use a (elimination-based) fixed-confidence best arm identification algorithm with $\delta=1/T$ to make a guess for the best arm and then commit to play this estimated best arm until the end of the horizon $T$. In a simple case (two Gaussian arms), we recently quantified the sub-optimality of such approaches: the regret of the best such Explore-Then-Commit strategy is at least twice larger than that of the kl-UCB algorithm~\cite{GKL16}. Even if the article focuses on Gaussian rewards, other cases with possibly more than two arms are also discussed. Coming back to the introductory example of A/B testing, the take-home message is the following: if you prefer to experiment first the two options before using only one of them in production, instead of continuously allocating the two options using a good regret-minimizing strategy, then this will cost you twice larger a regret. 
	
	Unlike the asymptotically optimal regret minimizing strategies that we presented, the asymptotically optimal Track-and-Stop strategy for best arm identification has no finite-time guarantees, and its implementation is slightly more complex. An important future work is to see whether useful tools for regret minimization, like the optimism principle or Bayesian methods can be combined with Track-and-Stop to have a simpler algorithm with a finite-time analysis. A starting point may be found in~\cite{Russo16}, who recently proposed a modified Bayesian Thompson Sampling rule that has some promising properties.
	
	\vspace{0.5cm}
	\noindent\textit{Aknowledgement:} The authors are extremely thankful to the reviewers of this paper, who contributed significantly to the clarity of the presentation by their numerous and always relevant comments. 
	
	\bibliographystyle{apalike}
	\bibliography{biblioBandits}

\begin{thebibliography}{}

\bibitem[Abbasi-Yadkori et~al., 2011]{YadLinear11}
Abbasi-Yadkori, Y., D.P{\'a}l, and C.Szepesv{\'a}ri (2011).
\newblock {Improved Algorithms for Linear Stochastic Bandits}.
\newblock In {\em {Advances in Neural Information Processing Systems}}.

\bibitem[Agrawal, 1995]{Agrawal:95}
Agrawal, R. (1995).
\newblock {Sample mean based index policies with O(log n) regret for the
  multi-armed bandit problem}.
\newblock {\em Advances in Applied Probability}, 27(4):1054--1078.

\bibitem[Agrawal et~al., 1989]{Agrawaletal89LBGene}
Agrawal, R., Teneketzis, D., and Anantharam, V. (1989).
\newblock {Asymptotically Efficient Adaptive Allocation Schemes for Controlled
  i.i.d. Processes: Finite Parameter Space}.
\newblock {\em IEEE Transactions on Automatic Control}, 34(3):258--267.

\bibitem[Agrawal and Goyal, 2012]{AGCOLT12}
Agrawal, S. and Goyal, N. (2012).
\newblock {Analysis of Thompson Sampling for the multi-armed bandit problem}.
\newblock In {\em {Proceedings of the 25th Conference On Learning Theory}}.

\bibitem[Agrawal and Goyal, 2013]{AGAISTAT13}
Agrawal, S. and Goyal, N. (2013).
\newblock {Further Optimal Regret Bounds for Thompson Sampling}.
\newblock In {\em {Proceedings of the 16th Conference on Artificial
  Intelligence and Statistics}}.

\bibitem[Audibert et~al., 2010]{Bubeck10BestArm}
Audibert, J.-Y., Bubeck, S., and Munos, R. (2010).
\newblock {Best Arm Identification in Multi-armed Bandits}.
\newblock In {\em {Proceedings of the 23rd Conference on Learning Theory}}.

\bibitem[Auer et~al., 2002]{Aueral02}
Auer, P., Cesa-Bianchi, N., and Fischer, P. (2002).
\newblock {Finite-time analysis of the multiarmed bandit problem}.
\newblock {\em Machine Learning}, 47(2):235--256.

\bibitem[Bechhofer, 1954]{Bechofer:54}
Bechhofer, R. (1954).
\newblock {A single-sample multiple decision procedure for ranking means of
  normal populations with known variances}.
\newblock {\em Annals of Mathematical Statistics}, 25:16--39.

\bibitem[Bechhofer et~al., 1968]{Bechofer:al68}
Bechhofer, R., Kiefer, J., and Sobel, M. (1968).
\newblock {\em {Sequential identification and ranking procedures}}.
\newblock The University of Chicago Press.

\bibitem[Bellman, 1956]{Bellman:Bay56}
Bellman, R. (1956).
\newblock {A problem in the sequential design of experiments}.
\newblock {\em The indian journal of statistics}, 16(3/4):221--229.

\bibitem[Berry and Fristedt, 1985]{Berry:Fristedt85}
Berry, D. and Fristedt, B. (1985).
\newblock {\em {Bandit Problems. Sequential allocation of experiments}}.
\newblock Chapman and Hall.

\bibitem[Bubeck, 2010]{Bubeck:Thesis}
Bubeck, S. (2010).
\newblock {\em {Jeux de bandits et fondation du clustering}}.
\newblock PhD thesis, Universit{\'e} de Lille 1.

\bibitem[Bubeck and Cesa-Bianchi, 2012]{Bubeck:Survey12}
Bubeck, S. and Cesa-Bianchi, N. (2012).
\newblock {Regret analysis of stochastic and nonstochastic multi-armed bandit
  problems}.
\newblock {\em Fondations and Trends in Machine Learning}, 5(1):1--122.

\bibitem[Bubeck et~al., 2013]{BG13}
Bubeck, S., Ernst, D., and Garivier, A. (Feb. 2013).
\newblock Optimal discovery with probabilistic expert advice: Finite time
  analysis and macroscopic optimality.
\newblock {\em Journal of Machine Learning Research}, 14:601--623.

\bibitem[Bubeck et~al., 2011]{Bubeckal11}
Bubeck, S., Munos, R., and Stoltz, G. (2011).
\newblock {Pure Exploration in Finitely Armed and Continuous Armed Bandits}.
\newblock {\em Theoretical Computer Science 412, 1832-1852}, 412:1832--1852.

\bibitem[Burnetas and Katehakis, 1996]{BurnKat96}
Burnetas, A. and Katehakis, M. (1996).
\newblock {Optimal adaptive policies for sequential allocation problems}.
\newblock {\em Advances in Applied Mathematics}, 17(2):122--142.

\bibitem[Capp{\'e} et~al., 2013]{KLUCBJournal}
Capp{\'e}, O., Garivier, A., Maillard, O.-A., Munos, R., and Stoltz, G. (2013).
\newblock {{K}ullback-{L}eibler upper confidence bounds for optimal sequential
  allocation}.
\newblock {\em Annals of Statistics}, 41(3):1516--1541.

\bibitem[Cesa-Bianchi and Lugosi, 2006]{PLG06}
Cesa-Bianchi, N. and Lugosi, G. (2006).
\newblock {\em {Prediction, Learning and Games}}.
\newblock Cambridge University Press.

\bibitem[Chapelle and Li, 2011]{LiChapelle11}
Chapelle, O. and Li, L. (2011).
\newblock {An empirical evaluation of Thompson Sampling}.
\newblock In {\em {Advances in Neural Information Processing Systems}}.

\bibitem[Chen et~al., 2016]{Chen16OptimalAlt}
Chen, L., Li, J., and Qiao, M. (2016).
\newblock Towards instance optimal bounds for best arm identification.
\newblock {\em arXiv:1608.06031}.

\bibitem[Chernoff, 1959]{Chernoff59}
Chernoff, H. (1959).
\newblock {Sequential design of Experiments}.
\newblock {\em The Annals of Mathematical Statistics}, 30(3):755--770.

\bibitem[Chu et~al., 2011]{LinUCB11}
Chu, W., Li, L., Reyzin, L., and Schapire, R. (2011).
\newblock {Contextual Bandits with Linear Payoff Functions}.
\newblock In {\em {Proceedings of the 14th Conference on Artificial
  Intelligence and Statistics}}.

\bibitem[Combes and Prouti{\`e}re, 2014]{Combes14Unimodal}
Combes, R. and Prouti{\`e}re, A. (2014).
\newblock {Unimodal Bandits without Smoothness}.
\newblock Technical report.

\bibitem[Even-Dar et~al., 2006]{EvenDaral06}
Even-Dar, E., Mannor, S., and Mansour, Y. (2006).
\newblock {Action Elimination and Stopping Conditions for the Multi-Armed
  Bandit and Reinforcement Learning Problems}.
\newblock {\em Journal of Machine Learning Research}, 7:1079--1105.

\bibitem[Gabillon et~al., 2012]{Gabillon:al12}
Gabillon, V., Ghavamzadeh, M., and Lazaric, A. (2012).
\newblock {Best Arm Identification: A Unified Approach to Fixed Budget and
  Fixed Confidence}.
\newblock In {\em {Advances in Neural Information Processing Systems}}.

\bibitem[Garivier and Capp{\'e}, 2011]{AOKLUCB}
Garivier, A. and Capp{\'e}, O. (2011).
\newblock {The {KL-UCB} algorithm for bounded stochastic bandits and beyond}.
\newblock In {\em {Proceedings of the 24th Conference on Learning Theory}}.

\bibitem[Garivier and Kaufmann, 2016]{GK16}
Garivier, A. and Kaufmann, E. (2016).
\newblock Optimal best arm identification with fixed confidence.
\newblock In {\em Proceedings of the 29th Conference On Learning Theory (to
  appear)}.

\bibitem[Garivier et~al., 2016a]{GKL16}
Garivier, A., Kaufmann, E., and Lattimore, T. (2016a).
\newblock On explore-then-commit strategies.
\newblock In {\em Advances in Neural Processing Systems (NIPS)}.

\bibitem[Garivier and Leonardi, 2011]{GL11}
Garivier, A. and Leonardi, F. (Nov. 2011).
\newblock Context tree selection: A unifying view.
\newblock {\em Stochastic Processes and their Applications},
  121(11):2488--2506.

\bibitem[Garivier et~al., 2016b]{GMS16}
Garivier, A., M{\'e}nard, P., and Stoltz, G. (2016b).
\newblock Explore first, exploit next: The true shape of regret in bandit
  problems.
\newblock {\em arXiv preprint arXiv:1602.07182}.

\bibitem[Garivier and Moulines, 2011]{GarivierMoulines11}
Garivier, A. and Moulines, E. (2011).
\newblock {On Upper-Confidence Bound Policies for Switching Bandit Problems}.
\newblock In {\em {Proceedings of the 22nd conference on Algorithmic Learning
  Theory}}.

\bibitem[Gittins, 1979]{Gittins79}
Gittins, J. (1979).
\newblock {Bandit processes and dynamic allocation indices}.
\newblock {\em Journal of the Royal Statistical Society, Series B},
  41(2):148--177.

\bibitem[Graves and Lai, 1997]{GravesLai97}
Graves, T. and Lai, T. (1997).
\newblock {Asymptotically Efficient adaptive choice of control laws in
  controlled markov chains}.
\newblock {\em SIAM Journal on Control and Optimization}, 35(3):715--743.

\bibitem[Jaksch et~al., 2010]{Auer:UCRL10}
Jaksch, T., Ortner, R., and Auer, P. (2010).
\newblock {Near-Optimal regret bounds for reinforcement learning}.
\newblock {\em Journal of Machine Learning Research}, 11:1563--1600.

\bibitem[Jamieson et~al., 2014]{Jamiesonal14LILUCB}
Jamieson, K., Malloy, M., Nowak, R., and Bubeck, S. (2014).
\newblock {lil'{UCB}: an Optimal Exploration Algorithm for Multi-Armed
  Bandits}.
\newblock In {\em {Proceedings of the 27th Conference on Learning Theory}}.

\bibitem[Kalyanakrishnan et~al., 2012]{Shivaramal12}
Kalyanakrishnan, S., Tewari, A., Auer, P., and Stone, P. (2012).
\newblock {{PAC} subset selection in stochastic multi-armed bandits}.
\newblock In {\em {International Conference on Machine Learning (ICML)}}.

\bibitem[Karnin et~al., 2013]{Karnin:al13}
Karnin, Z., Koren, T., and Somekh, O. (2013).
\newblock {Almost optimal Exploration in multi-armed bandits}.
\newblock In {\em {International Conference on Machine Learning (ICML)}}.

\bibitem[Katehakis and Robbins, 1995]{KatRob:95Gauss}
Katehakis, M. and Robbins, H. (1995).
\newblock {Sequential choice from several populations}.
\newblock {\em Proceedings of the National Academy of Science}, 92:8584--8585.

\bibitem[Kaufmann, 2014]{MaThese}
Kaufmann, E. (2014).
\newblock {\em {Analyse de strat{\'e}gies bay{\'e}siennes et fr{\'e}quentistes
  pour l'allocation s{\'e}quentielle de ressources}}.
\newblock PhD thesis.

\bibitem[Kaufmann, 2016]{AOS16}
Kaufmann, E. (2016).
\newblock On bayesian index policies for sequential resource allocation.
\newblock {\em Preprint arXiv:1601.01190}.

\bibitem[Kaufmann et~al., 2012a]{AISTATS12}
Kaufmann, E., Capp{\'e}, O., and Garivier, A. (2012a).
\newblock {On {B}ayesian {U}pper-{C}onfidence {B}ounds for Bandit Problems}.
\newblock In {\em {Proceedings of the 15th conference on Artificial
  Intelligence and Statistics}}.

\bibitem[Kaufmann et~al., 2014]{COLT14}
Kaufmann, E., Capp{\'e}, O., and Garivier, A. (2014).
\newblock {On the Complexity of A/B Testing}.
\newblock In {\em {Proceedings of the 27th Conference On Learning Theory}}.

\bibitem[Kaufmann et~al., 2016]{JMLR15}
Kaufmann, E., Capp{\'e}, O., and Garivier, A. (2016).
\newblock {On the Complexity of Best Arm Identification in Multi-Armed Bandit
  Models}.
\newblock {\em Journal of Machine Learning Research}, 17(1):1--42.

\bibitem[Kaufmann and Kalyanakrishnan, 2013]{COLT13}
Kaufmann, E. and Kalyanakrishnan, S. (2013).
\newblock {Information complexity in bandit subset selection}.
\newblock In {\em {Proceeding of the 26th Conference On Learning Theory.}}

\bibitem[Kaufmann et~al., 2012b]{ALT12}
Kaufmann, E., Korda, N., and Munos, R. (2012b).
\newblock {Thompson Sampling : an Asymptotically Optimal Finite-Time Analysis}.
\newblock In {\em {Proceedings of the 23rd conference on Algorithmic Learning
  Theory}}.

\bibitem[Lai and Robbins, 1985]{LaiRobbins85bandits}
Lai, T. and Robbins, H. (1985).
\newblock {Asymptotically efficient adaptive allocation rules}.
\newblock {\em Advances in Applied Mathematics}, 6(1):4--22.

\bibitem[Lattimore, 2016]{L16}
Lattimore, T. (2016).
\newblock Regret analysis of the anytime optimally confident {UCB} algorithm.
\newblock {\em CoRR}, abs/1603.08661.

\bibitem[Magureanu et~al., 2014]{Combes14Lip}
Magureanu, S., Combes, R., and Prouti{\`e}re, A. (2014).
\newblock {Lipschitz Bandits: Regret lower bounds and optimal algorithms}.
\newblock In {\em {Proceedings on the 27th Conference On Learning Theory}}.

\bibitem[Mannor and Tsitsiklis, 2004]{MannorTsi04}
Mannor, S. and Tsitsiklis, J. (2004).
\newblock {The Sample Complexity of Exploration in the Multi-Armed Bandit
  Problem}.
\newblock {\em Journal of Machine Learning Research}, pages 623--648.

\bibitem[M\'enard and Garivier, 2017]{MG17}
M\'enard, P. and Garivier, A. (Feb. 2017).
\newblock A minimax and asymptotically optimal algorithm for stochastic
  bandits.

\bibitem[Perchet and Rigollet, 2013]{PerchetRigollet13Covariates}
Perchet, V. and Rigollet, P. (2013).
\newblock The multi-armed bandit with covariates.
\newblock {\em The Annals of Statistics}.

\bibitem[Perchet et~al., 2015]{Perchet15Batched}
Perchet, V., Rigollet, P., Chassang, S., and Snowberg, E. (2015).
\newblock Batched bandit problems.
\newblock In {\em Proceedings of the 28th Conference On Learning Theory}.

\bibitem[Puterman, 1994]{Puterman94MDP}
Puterman, M. (1994).
\newblock {\em {Markov Decision Processes. Discrete Stochastic. Dynamic
  Programming.}}
\newblock Wiley.

\bibitem[Robbins, 1952]{Robbins52Freq}
Robbins, H. (1952).
\newblock {Some aspects of the sequential design of experiments}.
\newblock {\em Bulletin of the American Mathematical Society}, 58(5):527--535.

\bibitem[Russo, 2016]{Russo16}
Russo, D. (2016).
\newblock Simple bayesian algorithms for best arm identification.
\newblock In {\em Proceedings of the 29th Conference On Learning Theory}.

\bibitem[Scott, 2010]{Scott10}
Scott, S. (2010).
\newblock {A modern Bayesian look at the multi-armed bandit}.
\newblock {\em Applied Stochastic Models in Business and Industry},
  26:639--658.

\bibitem[Sutton and Barto, 1998]{SuttonBarto98}
Sutton, R. and Barto, A. (1998).
\newblock {\em Reinforcement Learning: an Introduction}.
\newblock MIT press.

\bibitem[Thompson, 1933]{Thompson33}
Thompson, W. (1933).
\newblock {On the likelihood that one unknown probability exceeds another in
  view of the evidence of two samples}.
\newblock {\em Biometrika}, 25:285--294.

\bibitem[Vaidhyan and Sundaresan, 2015]{Rajesh15Oddball}
Vaidhyan, N. and Sundaresan, R. (2015).
\newblock Learning to detect an oddball target.
\newblock {\em arXiv:1508.05572}.

\end{thebibliography}
	
\end{document}